\documentclass{article}

\usepackage[final,nonatbib]{nips_2018}

\usepackage[utf8]{inputenc} % allow utf-8 input
\usepackage[T1]{fontenc}    % use 8-bit T1 fonts
\usepackage{hyperref}       % hyperlinks
\usepackage{url}            % simple URL typesetting
\usepackage{booktabs}       % professional-quality tables
\usepackage{amsfonts}       % blackboard math symbols
\usepackage{nicefrac}       % compact symbols for 1/2, etc.
\usepackage{microtype}      % microtypography
\usepackage{color}
\usepackage{xcolor}
\usepackage{tcolorbox}
\usepackage{amsmath}
\usepackage{breqn}
\usepackage{float}
\usepackage{graphicx}
\usepackage{caption}
\usepackage{subcaption}
\usepackage{comment}
\usepackage{color}
\usepackage{etoolbox}
\usepackage{wrapfig}
\newbool{inccomment}
\setbool{inccomment}{true}

\usepackage{soul}
\usepackage{algorithmic}
\usepackage{graphicx}
\usepackage{textcomp}
\usepackage{xcolor}
\usepackage{ulem}
\usepackage[ruled]{algorithm2e}
\newcommand{\yan}[1]{\ifbool{inccomment}{{\color{blue}#1}}{}}

\newcommand{\das}[1]{\ifbool{inccomment}{{\color{red}#1}}{}}

\definecolor{dave-fancy}{RGB}{200,100,0}
\newcommand{\dave}[1]{\ifbool{inccomment}{{\color{dave-fancy}#1}}{}}
\newcommand{\davenote}[1]{\ifbool{inccomment}{{\color{pink}#1}}{}}

\title{Differentiable Fine-grained Quantization for Deep Neural Network Compression}
\author{
  Hsin-Pai Cheng\thanks{Equal Contribution and Co-First Authors}\\
  ECE Department\\
  Duke University\\
  Durham, NC 27708 \\
  \texttt{hc218@duke.edu}
  \And
  Yuanjun Huang\footnotemark[1]  \\
  University of Science \\
  and Technology of China\\
  Anhui, China\\
  \texttt{yjhuang@mail.ustc.edu.cn} \\
  \And
  Xuyang Guo\footnotemark[1] \\
  Tsinhua University\\
  Beijing, China\\
  \texttt{guoxuyang1997@gmail.com} \\
  \AND
  Feng Yan \\
  Computer Science\\ and Engineering\\
  University of Nevada, Reno\\
  Reno, NV 89557 \\
  \texttt{fyan@unr.edu} \\
  \And
  Yifei Huang \\
  Nanjing University\\
  Jiangsu, China\\
  \texttt{161240027@smail.nju.edu.cn} \\
  \And
  Wei Wen \\
  ECE Department\\
  Duke University\\
  Durham, NC 27708\\
  \texttt{wei.wen@duke.edu} \\
  \AND
  Hai Li \\
  ECE Department\\
  Duke University\\
  Durham, NC 27708 \\
  \texttt{hai.li@duke.edu} \\
  \And
  Yiran Chen \\
  ECE Department\\
  Duke University\\
  Durham, NC 27708 \\
  \texttt{yiran.chen@duke.edu} \\
}

\begin{document}

\maketitle

\begin{abstract}
%When deploying network models on mobile devices with limited resources, weight quantization has been widely adopted. Applying different precision in different layers/structures potentially can produce the most efficient model. In this work, we propose an automatic search algorithm based on relaxing the search space of quantization bitwidth from discrete to continuous domain, which demonstrated the ability to generate a mixed-precision quantization scheme whose compression rate is close to the one of binary-weighted model while the testing accuracy remains similar to the original full-precision model.
Neural networks have shown great performance in cognitive tasks. 
When deploying network models on mobile devices with limited resources, weight quantization has been widely adopted. 
Binary quantization obtains the highest compression but usually results in big accuracy drop.
In practice, 8-bit or 16-bit quantization is often used aiming at maintaining the same accuracy as the original 32-bit precision.  
%\dastin{
We observe different quantization schemes have different accuracy impact on different layers.
%~\cite{zhu2018adaptive}\cite{zhou2017adaptive}, so each layer has its own precision preference.} 
Thus judiciously selecting different precision for different layers/structures can potentially produce more efficient models compared to traditional quantization methods by striking a better balance between accuracy and compression rate.
%\yan{Seeking for the best precision configuration, however, is difficult due to the huge search space.}
In this work, we propose a fine-grained quantization approach for deep neural network compression by relaxing the search space of quantization bitwidth from discrete to a continuous domain.
%By relaxing the search space of quantization bitwidth from discrete to continuous domain, our algorithm can apply gradient descend based optimization approach and  generate a mixed-precision quantization scheme whose compression rate is close to the one of binary-weighted model while the testing accuracy remains similar to the original full-precision model.
The proposed approach applies gradient descend based optimization to generate a mixed-precision quantization scheme that outperforms the accuracy of traditional quantization methods under the same compression rate.

%\yanc{a general comment: shall we emphasis "fast" or "efficient" in addition to "automatic" or even replace "automatic"{} as exhaustive search can be considered as automatic?}{}
%\dastin{I think continuous relaxation's advantage over e{}xhaustive search lies in the efficiency. When the search space is too big, exhaustive search's algorithm complexity goes exponentially as the number of layer increases. However, our ongoing work on using LSTM-based relaxation approach can avoid such issue. But so far, the continuous relaxation approach discussed in this article still suffers from the search space problem--complexity goes exponentially as # of layer increases.}
\end{abstract}

\section{Introduction}
\label{sec:intro}
State-of-the-art neural networks have demonstrated promising performance in tasks such as image classification and object detection~\cite{googlenet}\cite{vgg}\cite{resnet}\cite{zfnet}. 
%\yanc{We may want to add more references here.}\dastin{Added} 
These network models are designed for high accuracy with less consideration in the computational cost and inference delay.
Thus deploying them 
%\begin{wrapfigure}{r}{0.34\textwidth}
\begin{figure}
    \centering
   %\vspace{-3mm}    
        \includegraphics[width=0.4\textwidth]{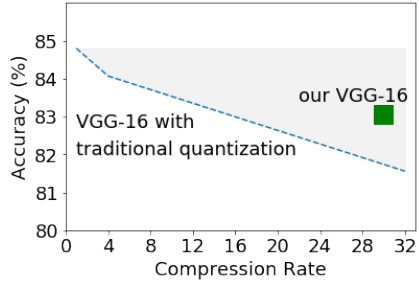}
     %   \vspace{-12pt}
        \caption{Accuracy VS compression rate. The dotted line is traditional quantization and the shaded area is our optimization goal. The box is our preliminary result.
        %where we achieve a higher accuracy in the same compression rate
        }
    \label{fig:grey_area}
  % \vspace{-6mm}
\end{figure}
%\end{wrapfigure}
on resource-constrained platform such as mobile phones is usually inefficient or even infeasible.
Even the recently proposed MobileNet~\cite{howard2017mobilenets}, which is customized for mobile platforms, has a relatively large 4.24M parameters. %(equivalent to approximately 16.96MB of memory). 
Extensive studies have been carried out in developing network models for resource constraint platforms.
Quantization is one of the most popular approaches~\cite{bnn}\cite{xnornet}. For example, Bai \textit{et al.} recently proposed a proximal operators for training low-precision deep neural netwokrs \cite{bai2018proxquant}. Courbariaux \textit{et al.} proposed a radical binary representation of the inputs, weights, and activations ~\cite{bnn}. Rastegari~\textit{et al.}~\cite{xnornet} theoretically analyzed the binary network and introduced a scaling scheme for their XNOR-Net---a network based on~\cite{bnn} with higher accuracy. Despite the significant improvement on inference accuracy, none of the above networks is able to achieve comparable accuracy as the full-precision counterparts. 
%\yanc{We may want to add some references here.}\dastin{Added}

\iffalse
For example, Courbariaux \textit{et al.}~\cite{bnn} and Rastegari \textit{et al.}~\cite{xnornet} incorporate binary quantization throughout the entire neural network,
while in practice, 8-bit or 16-bit quantization is more commonly adopted in order to minimize the accuracy drop. 
Dynamic fixed-point design with mixed precision was recently proposed by Mellempudi \textit{et al.}~\cite{mixpre}, which manually tuned the number of quantization clusters to trade-off the computational cost and accuracy.
\fi

%Quantization process indeed is a scaling operation which re-scales the floating-point weights in the original model to integer numbers.
%It is effective in reducing the weight bit-width of neural networks and the corresponding storage and computation requirement.
%The method is particularly useful and has been widely adopted in deploying network models on the platforms with limited computation, memory and power resources. 
 
%Our work differs from previous work in that we target searching the mixed-precision quantization scheme given a specific accuracy constraint.  and the feasibility of our algorithm is proven by its ability of dramatically shrinking the model size while maintaining an acceptable accuracy.
%This becomes the major obstacle that prevents binary neural network from being deployed in real-world applications.

We observe that different layers may have different accuracy sensitivity of quantization, thus a fine-grained quantization for each layer has the penitential to preserve accuracy under the same compression rate (defined as the ratio of original model size and compressed model size) compared to traditional course-grained quantization that uses the same quantization for the entire model.
%Maintaining some weights at high-bit precision is a potential a way to compensate for the accuracy loss brought by low-bit representation. 
%\dastin{Seems like we have to find another paper to cite?}For example, Mellempudi \textit{et al.}~\cite{mixpre} introduce the concept of mixed-precision quantization inference, which fixes the first layer's weight parameters to 8-bit integer and quantizes the remaining layers' parameters to another low bit representation.
%Our work focuses on searching for the optimized quantization bitwidth for each layer, which yields more flexibility compared with single-precision quantization for the whole model.
%More specifically, as shown in 
The dotted line in Figure~\ref{fig:grey_area} shows the trade-off between accuracy and compression rate in traditional quantization for VGG-16. 
Our goal is to push the trade-off between accuracy and compression rate into the shaded region of Figure~\ref{fig:grey_area} to achieve better compression efficiency, i.e., higher accuracy under the same compression rate. 
To achieve this, we propose a fine-grained quantization approach that relaxes the search space of quantization bitwidth from discrete to continuous domain and applies gradient descend optimization to generate best quantization scheme for each layer, i.e., applies lower bit for less quantization sensitive layers while preserving high bit precision for quantization sensitive layers.
Our experimental results show that the proposed approach outperforms the accuracy of traditional quantization methods under the same compression rate.

\section{Proposed Approach}
\label{sec:pro}
%Previous work showed that 8-bit quantization has little impact on the accuracy while still achieves $4\times$ compression compared to the original model with 32-bit floating-point weights.
%So it is intuitive to increase the quantization bitwidth of some weights to multiple bits to recover certain accuracy from binary neural networks.
%The approach was explored recently by ~\cite{mixpre}, in which the quantization clusters were selected manually. \yanc{need to double check as ~\cite{mixpre} uses clustering, which is also automatic.}\dastin{We will check that}
%Considering a large number of quantization combination across network layers, such a manual approach cannot be popularized to a general solution.

In this section, we propose a methodology to judiciously determine the best quantization scheme for each layer based on each layer's accuracy sensitivity of quantization. 
For easy description, we only use two-level quantization: binary and 8-bit quantization as an example to illustrate our approach and conduct a preliminary evaluation. It is straightforward to extend to more quantization levels.
%for each layer and therefore generate an optimal model.  
We relax the discrete variables to a continuous domain as the finer granularity of which can provide more accurate indication in quantization searching. %and boost the searc efficiency.
We adopt gradient descent based searching algorithm as it is fast and can be easily deployed in different machine learning frameworks.% implementation compatibility with most of the machine learning frameworks.
\begin{algorithm}[b]
\SetAlgoLined
 Initialization\;
 \While{not converged}{
  Update weights ${w}$ by descending $\nabla_{w}\mathcal{L}_{train}({w,\alpha})$\;
  \eIf{$\mathcal{L}_{valid}({w,\alpha})-\theta\leq0$}{
   $\lambda\gets0$\;
   }{
   $\lambda\gets{inf}$\;
  }
  Update probability ${\alpha}$ by descending $\nabla_{\alpha}(\mathcal{G}(\alpha)+\lambda(\mathcal{L}_{valid}({w^{*},\alpha})-\theta))$\;
 }
\caption{Differentiable fine-grained quantization}
\end{algorithm}

We use \textit{Softmax} function to relax the search space from discrete to continuous.
We denote the output of layer $i$ with continuous relaxation as $q_i$. For example, binary and 8-bit quantization represented as $q_{i_0}$ and $q_{i_1}$. 
\textit{Softmax}$(q_{i_0},q_{i_1})$ can be translated as the probability of binary and 8-bit quantization, respectively.
Thus $q_i$ can be computed as

\begin{equation}
    q_{i}=\frac{\sum_{j=0}^1exp(\alpha_{i_j})\mathcal{B}(q_{i_j})}{\sum_{j=0}^1exp(\alpha_{i_j})},
    \label{eq:1}
\end{equation}
where $\mathcal{B}$ is the batch normalization operation. 
The output $q_{i}$ is used as the input of the following layer. 
The search space for a network with $k$ layer is $\alpha=\{\alpha_{0},\alpha_{1},..., \alpha_{k-1}\}$. 
To explore the trade-off between different quantization schemes, we model the target objective function as
\begin{equation}
    \min\limits_{\alpha}\mathcal{G}(\alpha),
    \label{con:1}
\end{equation}
\begin{equation}
    s.t.~~~~\mathcal{L}_{val}({w^{*},\alpha})- \theta\leq0,
    \label{con:2}
\end{equation}
where
\begin{equation}
    {w^{*}}=\arg\min\limits_{w}\mathcal{L}_{train}({w^{*},\alpha}).
    \label{con:3}
\end{equation}
Here $\mathcal{G}$ represents the model size, 
$\mathcal{L}$ is the cross entropy loss, 
$\theta$ is the expected maximum loss, 
$w$ denotes the weights of the model,
and $\alpha$ represents the coefficient of either quantization method (binary or 8-bit) in a certain layer.
In our model, (\ref{con:2}) is the constraint for optimization problem (\ref{con:1}). We can rewrite the above as a bi-level optimization problem: 
\begin{equation}
\begin{aligned}
    \min\limits_{\alpha}\max\limits_{\lambda\geq0} & \left(\mathcal{G}(\alpha)+\lambda(\mathcal{L}(\alpha,w^{*})-\theta)\right)\\ 
    s.t. & \qquad w^{*}=\arg\min\limits_{w}\mathcal{L}_{train}(w,\alpha).
\end{aligned}
\label{prob:1}
\end{equation}
To solve this bi-level optimization problem, we adopt the approximate algorithm in~\cite{liu2018darts}. 
%\dastin{
First, we retrain the network to find the weights that result in the minimal loss on the training set.
%and to make sure that the model is stable at the current learned value of alpha and weights. }
Then the Lagrange multiplier problem is solved by fixing the weights.
As shown in Algorithm 1, solving the Lagrange multiplier problem starts with maximizing the target function w.r.t. $\lambda$: 
if $\mathcal{L}(\alpha,w^{*})-\theta \leq0$, $\lambda$ approaches 0; otherwise $\lambda$ approaches infinite. Here $\theta$ is a tunable hyperparameter representing the tolerance of accuracy drop. Larger $\theta$ tolerates less accuracy drop but may also result in smaller compression rate. While smaller $\theta$ can potentially achieve a higher compression rate, it may cause larger accuracy drop. Our setting of $\theta$ is using the (expected or target) loss in full precision model. 
Finally we minimize the target function w.r.t. $\alpha$.
Once obtaining the hyperparameter set $\alpha$ with the best trade-off, we retrain the quantization and fine tune the quantized weights to generate the final network model.

%\yanc{need to add an analysis about search complexity, e.g., from originally exponential to polynomial.}
%\dastin{I think our complexity is still expoential}

%\hal{Thus in our target problem, condition (\ref{con:3}) is the constraint of condition (\ref{con:2}) whereas condition (\ref{con:2}) is the constraint of condition (\ref{con:1}). Considering condition (\ref{con:2}) and (\ref{con:1}) as the problem which can be solved by Lagrange multiplier method, (\ref{con:3}) is the constraint of the new problem. [Rewrite this paragraph.]}

%After finding the hyperparameter set $\alpha$ with the best trade-off, we retrain the quantization operation with \hal{the strongest $\alpha$ [what is strongest?]} at every layer. 
%After retaining the quantization operation and fine tuning the quantized weights we will have the final solution for the inference.

\section{Experimental Evaluation}
%We tested our method on a pretrained 2-layer depthwise separable convolution network for MNIST dataset as well as a more complicated VGG-16 model for CIFAR-10 dataset.For each model, we compare our method with 32-bit floating-point (full-precision baseline), 8-bit fixed-point and binary neural networks. The results are summarized in Table~1. In MNIST experiment, our algorithm is capable of searching a scheme which achieves 28x compression rate while keeping accuracy drop less than 0.5\%. The CIFAR-10 experiment shows a compression rate much close to that of binary representation with 1.5\% higher accuracy.

We evaluate our proposed methodology on a pretrained 2-layer depth-wise separable convolution neural network using MNIST data set as well as VGG-16 neural network model using CIFAR-10 data set. For each model, we compare our approach with the following baselines: 32-bit floating point (full-precision) model, 8-bit fixed precision model, and binary fixed precision model. As shown in Table \ref{tab:result}, the results of MNIST experiment suggest that our algorithm is capable of find a quantization scheme that achieves 28x compression rate while keeping the accuracy drop less than 0.5\%. 
In CIFAR-10 experiment, we set $\theta$ as 0.6 for VGG-16. Comparing to whole binary quantization, our approach obtains a compression rate that is very close to binary quantization while gaining 1.5\% more accuracy. 
%Here the compression rate is calculated by original model size divided by compressed model size. %Our quantization approach can also be combined with network pruning ~\cite{zhong2018prune}, which is reported to be capable of pruning 84.7 \% of FLOPS on a VGG-19 network. Thus, our approach theoretically has the potential of compressing the VGG-16 network by about 40x altogether.
Figure 2 shows the memory consumption of our model and the original 32-bit full precision model. The memory usage is dramatically decreased especially at the middle layers. 
It is worth mentioning that our method is orthogonal to weight pruning. Combing with state-of-the-art pruning methods~\cite{zhang2018adam}\cite{progadmm} which achieve approximately 30x compression rate, the overall compression rate can be up to approximately 900x.

\begin{figure}[t]
     
   % \vspace{-6pt}
\end{figure}

\ifx % commented by Dave
\begin{figure}[h]
    \centering
        \includegraphics[width=\columnwidth]{acc_loss1.png}
        \caption{Accuracy and Loss w.r.t. iteration}
    \label{fig:acc&loss}
\end{figure}
\fi

\ifx % commented by Dave
\begin{figure}[h]
    \centering
        \includegraphics[width=0.5\textwidth]{layers1.png}
        \caption{This figure illustrates how the probability of binary quantizaion in certain layers trend as iteration increases. All probabilities are initialized to 0.5.}
    \label{fig:prob}
\end{figure}
\fi

% \usepackage{booktabs}

% \begin{table}[t]
% \caption{Comparison of different quantization schemes.}
% \label{tab:result}
% \centering
% \small
% \begin{tabular}{p{65pt}|p{30pt}|p{30pt}|p{30pt}|p{65pt}|p{30pt}|p{30pt}|p{30pt} } 
% \hline
% Model / Data & Quant. & Comp. & Accu. & Model / Data & Quant. & Comp. & Accu. \\
% \hline
% Depthwise Sep. / & float32 & 1$\times$ & 98.66\% & VGG16 / & float32 & 1$\times$ & 84.80\%\\
% Conv. & 8-bit & 4$\times$ & 98.48\% & CIFAR-10 & 8-bit & 4$\times$ & 84.07\%\\ 
% & \textbf{ours} & \textbf{28$\times$} & \textbf{98.2\%} & & \textbf{ours} & \textbf{30$\times$} & \textbf{83.06\%}\\
% & binary & 32$\times$ & 96.34\% & & binary & 32x & 81.56\%\\
% \hline
% \end{tabular}

% \vspace{-12pt}
% \end{table}

\begin{table}
  \begin{minipage}[t!]{0.5\linewidth}
  \caption{Comparison of different quantization schemes.}
  \label{tab:result}
  \centering
  \begin{tabular}{cccccccc}
   %\begin{tabular}{p{65pt}|p{30pt}|p{30pt}|p{30pt}|p{65pt}|p{30pt}|p{30pt}|p{30pt} }
    \toprule
     &\multicolumn{2}{c}{MNIST} &\multicolumn{2}{c}{CIFAR-10} \\
    \cmidrule(r){2-5}
    
   Quant.  & Comp.  & Accu.(\%) & Comp.  & Accu.(\%) \\
    \midrule
    float32 & 1$\times$ & 98.66 & 1$\times$  & 84.80 \\
    \midrule
    8-bit   & 4$\times$ & 98.48 & 4$\times$ & 84.07\\
    \midrule
    \textbf{ours} & \textbf{28}$\times$ & \textbf{98.20} &  \textbf{30}$\times$  & \textbf{83.06} \\
    \midrule
     binary & 32$\times$ & 96.34 & 32 & 81.56 \\
    \bottomrule
  \end{tabular}
  \end{minipage}\hfill
  \begin{minipage}[t!]{0.41\linewidth}
  
  \centering
        \includegraphics[width=40mm]{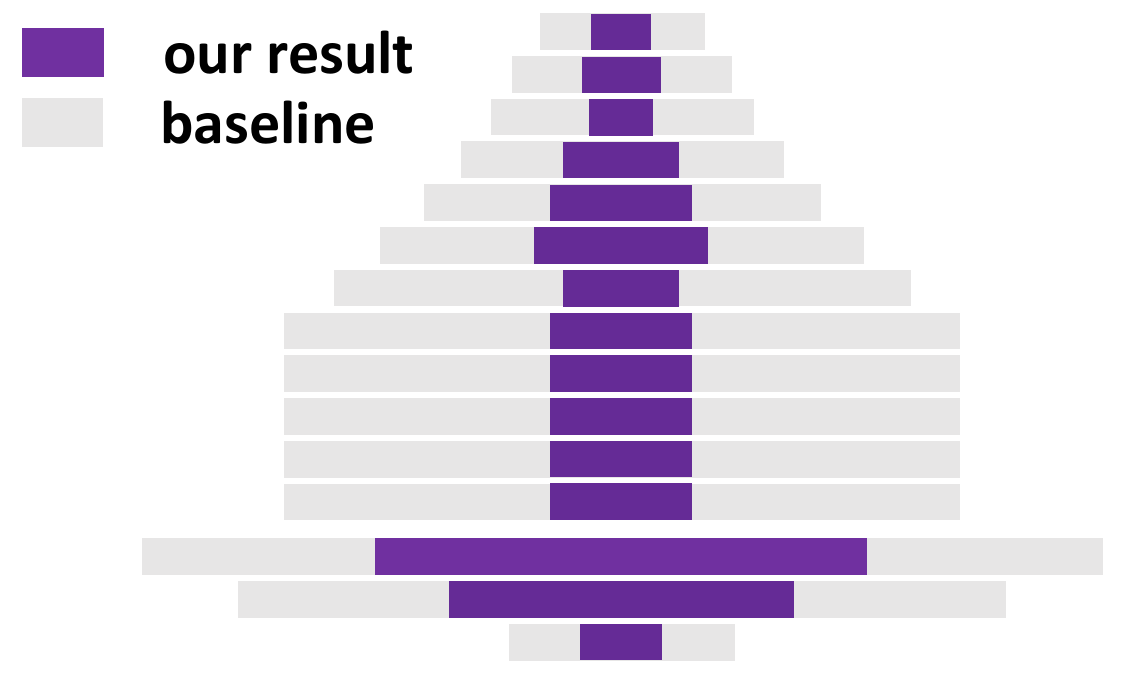}
        \vspace{3mm}
        \captionof{figure}{Pretrained 32 bit VGG-16 vs. the mixed precision model generated by our algorithm. 
        The width of a rectangle denotes the size (i.e., memory consumption) of the corresponding layer.
        }
    \label{fig:solution}
  \end{minipage}
\end{table}
\section{Conclusion and On-going Work}
%We explored the idea of mixed-precision quantization on pretrained model and proposed an approach to conduct continuous relaxation of the quantization bitwidths in discrete domain. Our method proved that the trade-off between computational resources and model accuracy can be achieved via an automatic search algorithm. More specifically, the evaluation on MNIST and CIFAR-10 datasets shows the network model generated by our method remains the similar accuracy to the full-precision model while its compression rate is close to the binary counterpart. Given the similarity of weight quantization problem in different kinds of neural topologies, we think that our work can potentially be extended to mixed-precision quantization of RNNs, which are commonly used in natural language processing.
%\yanc{we can add that this work can be potentially extended to RNN/LSTM etc..}
%\dastin{Added}

%We plan to continue the research on shortening the search process by pruning unnecessary search space in the future.

In this paper, we propose a differentiable mixed-precision search method for compressing deep neural networks efficiently. Unlike the traditional quantization methods, our approach relaxes quantization bitwidths to a continuous domain and combined with loss function. Deep neural networks can be either quantized from the start of training phase or from a pretrained model using our proposed methodology. Moreover, our approach ensures quantized model remain a similar accuracy while being compressed up to 30X. 
%Given the similarity of weight quantization problem in different kinds of neural network topologies, our proposed 

The proposed methodology is not tied into any specific neural network topology, so it can potentially be extended to mixed-precision quantization of different neural network architectures, such as  RNN and LSTM.
We are currently working on providing more quantization options for each layer. For example, each layer can be quantized to $x$ bits, and $x \in \{1, 2, ..., 32\} $. These new quantization options drastically increase the search space. Therefore, We plan to design a predictor combined with autoencdoer-decoder architecture to expedite the search process of layer-wise quantization.

%\hl{Dave: I moved the algorithm to bottom, rewrote ongoing work, conclusion. Ongoing work is combined with conclusion. I also fixed some words at experiment}

\section*{Acknowledgement}

This work is supported in part by the following grants: National Science Foundation CCF-1756013, IIS-1838024, 1717657 and Air Force Research Laboratory FA8750-18-2-0057. 

\medskip

\small

\bibliographystyle{unsrt}
\bibliography{egbib.bib}

\end{document}